\documentclass[10pt,twocolumn,letterpaper]{article}

\usepackage{cvpr}
\usepackage{times}
\usepackage{epsfig}
\usepackage{graphicx}
\usepackage{amsmath}
\usepackage{amssymb}

\usepackage{color}
\usepackage{diagbox}
\usepackage[caption=false,font=scriptsize]{subfig}
\usepackage[linesnumbered,lined,boxed,commentsnumbered]{algorithm2e}
\usepackage{multirow}

\cvprfinalcopy % *** Uncomment this line for the final submission

\def\cvprPaperID{****} % *** Enter the CVPR Paper ID here

\ifcvprfinal\fi
\begin{document}

%%%%%%%%% TITLE
\title{Wing Loss for Robust Facial Landmark Localisation with Convolutional Neural Networks}

\author{Zhen-Hua Feng$^1$  ~~Josef Kittler$^1$  ~~Muhammad Awais$^1$  ~~Patrik Huber$^1$  ~~Xiao-Jun Wu$^2$\\
$^1$ Centre for Vision, Speech and Signal Processing, University of Surrey, Guildford GU2 7XH, UK\\
$^2$School of IoT Engineering, Jiangnan University, Wuxi 214122, China\\
{\tt\small \{z.feng, j.kittler, m.a.rana\}@surrey.ac.uk, patrikhuber@gmail.com, wu\_xiaojun@jiangnan.edu.cn}
% For a paper whose authors are all at the same institution,
% omit the following lines up until the closing ``}''.
% Additional authors and addresses can be added with ``\and'',
% just like the second author.
% To save space, use either the email address or home page, not both
}

\maketitle
\thispagestyle{empty}

%%%%%%%%% ABSTRACT
\begin{abstract}
We present a new loss function, namely Wing loss, for robust facial landmark localisation with Convolutional Neural Networks (CNNs). We first compare and analyse different loss functions including L2, L1 and smooth L1. The analysis of these loss functions suggests that, for the training of a CNN-based localisation model, more attention should be paid to small and medium range errors. To this end, we design a piece-wise loss function. The new loss amplifies the impact of errors from the interval (-w, w) by switching from L1 loss to a modified logarithm function.

To address the problem of under-representation of samples with large out-of-plane head rotations in the training set, we propose a simple but effective boosting strategy, referred to as pose-based data balancing. In particular, we deal with the data imbalance problem by duplicating the minority training samples and perturbing them by injecting random image rotation, bounding box translation and other data augmentation approaches. Last, the proposed approach is extended to create a two-stage framework for robust facial landmark localisation. The experimental results obtained on AFLW and 300W demonstrate the merits of the Wing loss function, and prove the superiority of the proposed method over the state-of-the-art approaches.
%Our model is available at \url{https://github.com/FengZhenhua/Wing-Loss}.
\end{abstract}

\section{Introduction}
Facial landmark localisation, or face alignment, aims at finding the coordinates of a set of pre-defined key points for 2D face images.
A facial landmark usually has specific semantic meaning, \eg nose tip or eye centre, which provides rich geometric information for other face analysis tasks such as face recognition~\cite{Taigman2014, Masi_2016_CVPR, Liu_2017_CVPR, Yang_2017_CVPR}, emotion estimation~\cite{Zeng_2009_PAMI,Benitez_Quiroz_2016_CVPR, Walecki_2016_CVPR, Li_2017_CVPR} and 3D face reconstruction~\cite{Dou_2017_CVPR,kittler20163d, huber_2017_spl,hu2017efficient, Roth_2016_CVPR, koppen2018gaussian,feng2018evaluation}.
\begin{figure}[!t]
\centering
\subfloat[$w=5$]{
 \label{fig_wing_1}
\includegraphics[trim = 61mm 100mm 62mm 100mm, clip, width = .49\linewidth]{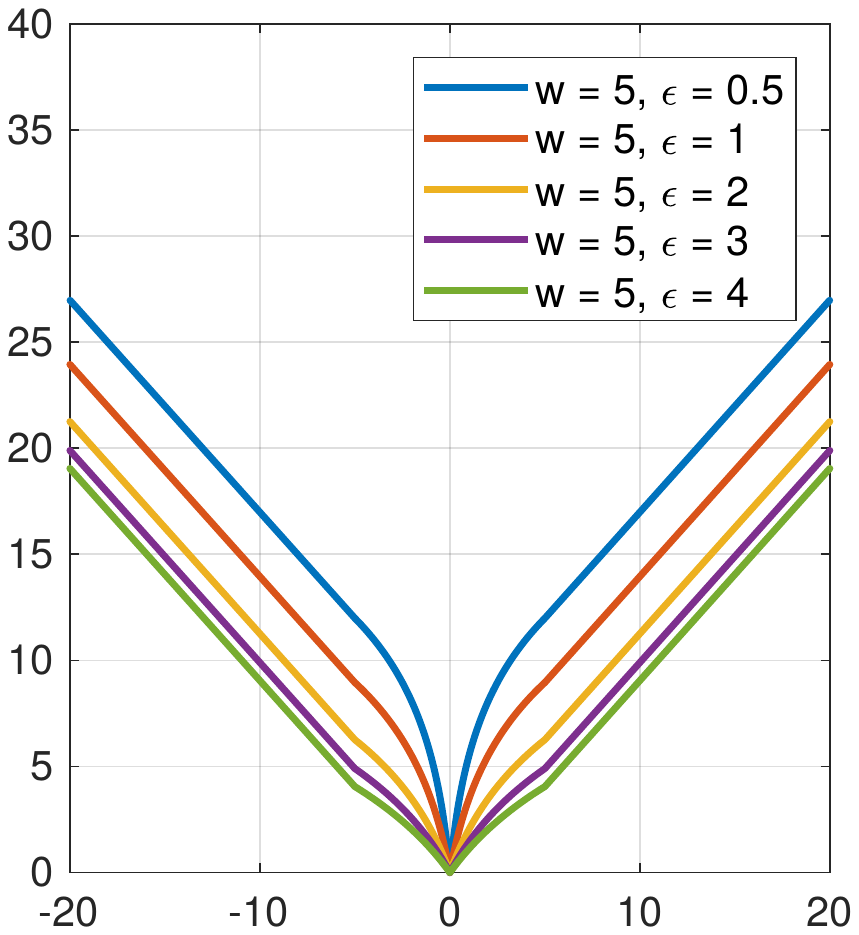}
}
\subfloat[$w=10$]{
 \label{fig_wing2}
 \includegraphics[trim = 61mm 100mm 62mm 100mm, clip, width = .49\linewidth]{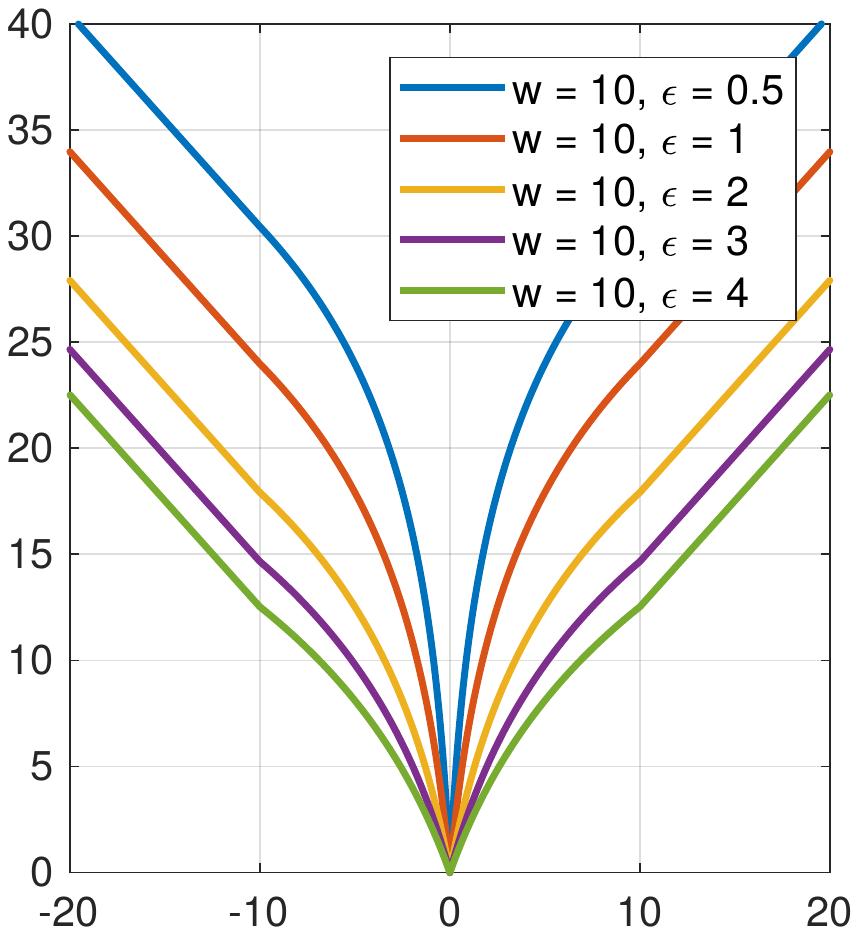}
}
\caption{Our Wing loss function (Eq.~\ref{equ_wing_loss}) plotted with different parameter settings, where $w$ limits the range of the non-linear part and $\epsilon$ controls the curvature. By design, we amplify the impact of the samples with small and medium range errors to the network training.}
\label{fig_wing}
\end{figure}

Thanks to the successive developments in this area of research during the past decades, we are able to perform very accurate facial landmark localisation in constrained scenarios, even using traditional approaches such as Active Shape Model (ASM)~\cite{Cootes1995}, Active Appearance Model (AAM)~\cite{Cootes2001} and Constrained Local Model (CLM)~\cite{Cristinacce2006}.
The existing challenge is to achieve robust and accurate landmark localisation of unconstrained faces that are impacted by a variety of appearance variations, \eg in pose, expression, illumination, image blurring and occlusion.
To this end, cascaded-regression-based approaches have been widely used, in which a set of weak regressors are cascaded to form a strong regressor~\cite{dollar2010cascaded, xiong2013supervised, Cao2014face, FENG2015, Wu_2016_CVPR, Wu_2017_CVPR, feng2017face}.
However, the capability of cascaded regression is nearly saturated due to its shallow structure.
After cascading more than four or five weak regressors, the performance of cascaded regression is hard to improve further~\cite{sun2015cascaded, feng2015cascaded}.
More recently, deep neural networks have been put forward as a more powerful alternative in a wide range of computer vision and pattern recognition tasks, including facial landmark localisation~\cite{sun2013deep, zhang2016joint, Zhang_2016_CVPR, Lv_2017_CVPR, yang2017stacked, wu2017facial, ranjan2017hyperface}.

To perform robust facial landmark localisation using deep neural networks, different network types have been explored, such as the Convolutional Neural Network (CNN)~\cite{sun2013deep}, Auto-Encoder Network~\cite{zhang2014coarse} and Recurrent Neural Network (RNN)~\cite{Trigeorgis_2016_CVPR, Xiao2016}.
In addition, different network architectures have been extensively studied during the recent years along with the development of deep neural networks in other AI applications.
For example, the Fully Convolutional Network (FCN)~\cite{liang2015unconstrained} and hourglass network with residual blocks have been found very effective~\cite{newell2016stacked, yang2017stacked, deng2017joint}.

One crucial aspect of deep learning is to define a loss function leading to better-learnt representation from underlying data. 
However, this aspect of the design seems to be little investigated by the facial landmark localisation community.
% As deep-neural-network-based facial landmark localisation is learning-based and data-driven, the design of a proper loss function is crucial for successful network training.
% However, this aspect of the design  seems to have been overlooked by the community.
To the best of our knowledge, most existing facial landmark localisation approaches using deep learning are based on the L2 loss.
However, the L2 loss function is sensitive to outliers, which has been noted in connection with the bounding box regression problem in the well-known Fast R-CNN algorithm~\cite{girshick2015fast}.
Rashid \etal also notice this issue and use the smooth L1 loss instead of L2~\cite{Rashid_2017_CVPR}.
To further address the issue, we propose a new loss function, namely Wing loss (Fig.~\ref{fig_wing}), for robust facial landmark localisation.
The main contributions of our work include:
\begin{itemize}
    \item presenting a systematic analysis of different loss functions that could be used for regression-based facial landmark localisation with CNNs, which to our best knowledge is the first such study carried out in connection with the landmark localisation problem.
    We empirically and theoretically compare L1, L2 and smooth L1 loss functions and find that L1 and smooth L1 perform much better than the widely used L2 loss.
    
    \item a novel loss function, namely the Wing loss, which is designed to improve the deep neural network training capability for small and medium range errors.
    
    \item a data augmentation strategy, \ie pose-based data balancing, that compensates the low frequency of occurrence of samples with large out-of-plane head rotations in the training set.
    
    \item a two-stage facial landmark localisation framework for performance boosting.
\end{itemize}

The paper is organised as follows. 
Section~\ref{sec_2} presents a brief review of the related literature. The regression-based facial landmarking problem with CNNs is formulated in Section~\ref{sec_3}. 
The properties of common loss functions (L1 and L2) are discussed in Section~\ref{sec_4} which also motivate the introduction of the novel Wing loss function. 
The pose-based data balancing strategy is the subject of Section~\ref{sec_5}. 
The two-stage localisation framework is proposed in Section~\ref{sec_6}.
The advocated approach is validated experimentally in Section~\ref{sec_7} and the paper is drawn to conclusion in Section~\ref{sec_8}.
%%% Related work
%%%------------------------------------------------------------
%%%------------------------------------------------------------
%%%------------------------------------------------------------
%%%------------------------------------------------------------
%%%------------------------------------------------------------
%%%------------------------------------------------------------
\section{Related work}
\label{sec_2}
\textbf{Network Architectures:}
Most deep-learning-based facial landmark localisation approaches are regression-based.
For such a task, the most straightforward way is to use a CNN model with regression output layers~\cite{sun2013deep, Rashid_2017_CVPR}.
The input for a regression CNN is usually an image patch enclosing the whole face region and the output is a vector consisting of the 2D coordinates of facial landmarks.
Besides the classical CNN architecture, newly developed CNN systems have also been used for facial landmark localisation and shown promising results, \eg FCN~\cite{liang2015unconstrained} and the hourglass network~\cite{newell2016stacked, yang2017stacked, deng2017joint,Bulat_2017_ICCV, Bulat_2017_ICCV_2}.
Different from traditional CNN-based approaches, FCN and hourglass network output a heat map for each landmark.
These heat maps are of the same size as the input image.
The value of a pixel in a heat map indicates the probability that its location is the predicted position of the corresponding landmark.
To reduce false alarms of a generated 2D sparse heat map, Wu \etal propose a distance-aware softmax function that facilitates the training of their dual-path network~\cite{wu2017godp}.

Thanks to the extensive studies of different deep neural networks and their use cases in unconstrained facial landmark localisation, the development of the area has been  greatly promoted. 
However, the current research lacks a systematic analysis on the use of different loss functions.
In this paper, we close this gap and design a new loss function for CNN-based facial landmark localisation.

\textbf{Dealing with Pose Variations:}
Extreme pose variations bring many difficulties to unconstrained facial landmark localisation.
To mitigate this issue, different strategies have been explored.
The \textit{first} one is to use multi-view models.
There is a long history of the use of multi-view models in landmark localisation, from the earlier studies on ASM~\cite{Romdhani1999} and AAM~\cite{cootes2002view} to recent work on cascaded-regression-based~\cite{xiong2015global, zhu2016unconstrained, Feng_2017_CVPR} and deep-learning-based approaches~\cite{deng2017joint}.
For example, Feng \etal train multi-view cascaded regression models using a fuzzy membership weighting strategy, which, interestingly, outperforms even some deep-learning-based approaches~\cite{Feng_2017_CVPR}.
The \textit{second} strategy, which has become very popular in recent years, is to use 3D face models~\cite{zhu2016face, Jourabloo_2016_CVPR, Bhagavatula_2017_ICCV, Liu_2017_ICCV_Workshops, Jourabloo_2017_ICCV}.
By recovering the 3D shape and estimating the pose of a given input 2D face image, the issue of extreme pose variations can be alleviated to a great extent.
In addition, 3D face models have also been widely used to synthesise additional 2D face images with pose variations for the training of a pose-invariant system~\cite{masi2016we,feng2015cascaded,zhu2016face}.
\textit{Last}, multi-task learning has been adopted to address the difficulties posed by image degradation, including pose variations.
For example, face attribute estimation, pose estimation or 3D face reconstruction can jointly be trained with facial landmark localisation~\cite{zhang2016joint, Xu2017joint, ranjan2017hyperface}.
The collaboration of different tasks in a multi-task learning framework can boost the performance of individual sub-tasks.

Different from these approaches, we treat the challenge as a training data imbalance problem and advocate a pose-based data balancing strategy to address this issue.

\textbf{Cascaded Networks:}
In the light of the coarse-to-fine cascaded regression framework, multiple networks can be stacked to form a stronger network to boost the performance.
To this end, shape- or landmark-related features should be used to satisfy the training of multiple networks in cascade.
However, a CNN using a global face image as input cannot meet this requirement.
To address this issue, one solution is to extract CNN features from local patches around facial landmarks.
This idea is advocated, for example, by Trigeorgis \etal who use the Recurrent Neural Network (RNN) for end-to-end model training~\cite{Trigeorgis_2016_CVPR}.
As an alternative, we can train a network based on the global image patch for rough facial landmark localisation.
Then, for each landmark or a composition of multiple landmarks in a specific region of the face, a network is trained to perform fine-grained landmark prediction~\cite{Sun2013, dong2015adaptive, Lv_2017_CVPR, Xu2017joint}.
For another example, Yu \etal propose to inject local deformations to the estimated facial landmarks of the first network using thin-plate spline transformations~\cite{yu2016deep}.

In this paper, we use a two-stage CNN-based landmark localisation framework. 
The first CNN is a very simple one that can perform rough facial landmark localisation very quickly.
The aim of the first network is to mitigate the difficulties posed by inaccurate face detection and in-plane head rotations.
Then the second CNN is used to perform fine-grained landmark localisation.

%%%------------------------------------------------------------
%%%------------------------------------------------------------
%%%------------------------------------------------------------
%%%------------------------------------------------------------
%%%------------------------------------------------------------
%%%------------------------------------------------------------
\section{CNN-based facial landmark localisation}
\label{sec_3}
The target of CNN-based facial landmark localisation is to find a nonlinear mapping:
\begin{equation}
    \Phi: \mathcal{I} \rightarrow \mathbf{s},
\end{equation} 
that outputs a shape vector $\mathbf{s} \in \mathbb{R}^{2L}$ for a given input colour image $\mathcal{I} \in \mathbb{R}^{H \times W \times 3}$.
The input image is usually cropped using the bounding box output by a face detector.
The shape vector is in the form of $\mathbf{s} = [x_1, ..., x_L, y_1, ..., y_L]^T$, where $L$ is the number of pre-defined 2D facial landmarks and $(x_l, y_l)$ are the coordinates of the $l$th landmark.
To obtain this mapping, first, we have to define the architecture of a multi-layer neural network with randomly initialised parameters.
In fact, the mapping $\Phi = (\phi_1 \circ ... \circ \phi_M)(\mathcal{I})$ is a composition of $M$ functions, in which each function stands for a specific layer in the network.
\begin{figure}[t]
\centering
   \includegraphics[trim={0mm 106mm 93mm 0mm}, clip, width=1\linewidth]{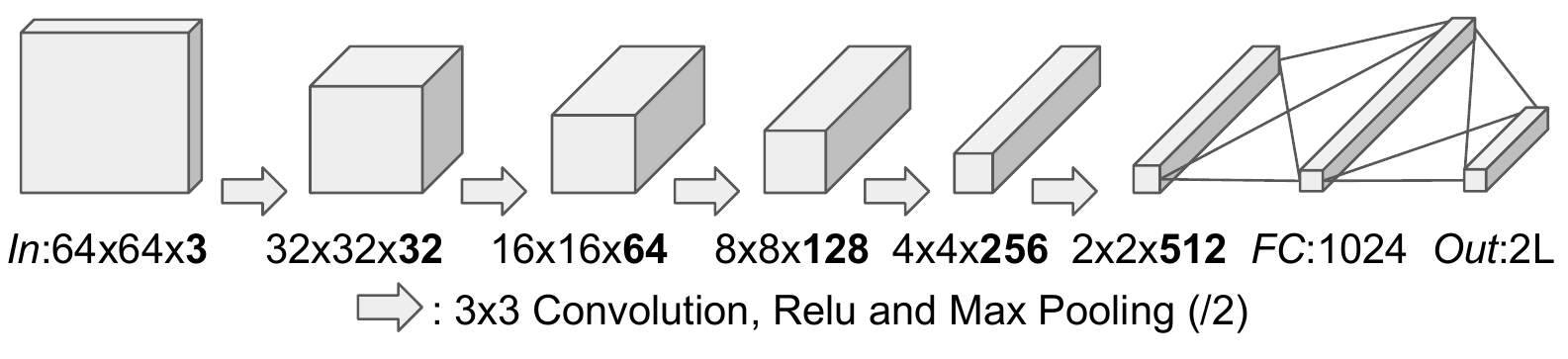}
   \caption{Our simple CNN-6 network consisting of 5 convolutional and 1 fully connected layers followed by an output layer.}
\label{fig_cnn64}
\end{figure}

Given a set of labelled training samples $\Omega = \{\mathcal{I}_i, \mathbf{s}_i\}_{i=1}^{N}$, the target of CNN training is to find a $\Phi$ that minimises:
\begin{equation}
\label{equ_obj}
    \sum_{i=1}^{N} loss(\Phi(\mathcal{I}_i), \mathbf{s}_i),
\end{equation} 
where $loss()$ is a pre-defined loss function that measures the difference between a predicted shape vector and its ground truth.
In such a case, the CNN is used as a regression model learned in a supervised manner.
To optimise the above objective function, optimisation algorithms such as Stochastic Gradient Descent (SGD) can be used.

To empirically analyse different loss functions, we use a simple CNN architecture, in the following termed CNN-6, for facial landmark localisation, to achieve high speed in model training and testing.
% The input for this network is a $64 \times 64 \times 3$ colour image hence we use the term `CNN-64' for this network.
The input for this network is a $64 \times 64 \times 3$ colour image and the output is a vector of $2L$ real numbers for the 2D coordinates of $L$ landmarks.
As shown in Fig.~\ref{fig_cnn64}, our CNN-6 has five $3\times3$ convolutional layers, a fully connected layer and an output layer.
After each convolutional and fully connected layer, a standard Relu layer is used for nonlinear activation. 
A Max pooling after each convolutional layer is used to downsize the feature map to half of the size.

To boost the performance, more powerful network architectures can be used, such as our two-stage landmark localisation framework presented in Section~\ref{sec_6} and the recently proposed ResNet architecture~\cite{he2016deep}.
We will report the results of these advanced network architectures in Section~\ref{sec_7}.
It should be highlighted that, to the best of our knowledge, this is the first time that such a deep residual network, \ie ResNet-50, is used for facial landmark localisation.

%%% Analysis of loss functions
%%%------------------------------------------------------------
%%%------------------------------------------------------------
%%%------------------------------------------------------------
%%%------------------------------------------------------------
%%%------------------------------------------------------------
%%%------------------------------------------------------------
\section{Wing loss}
\label{sec_4}
The design of a proper loss function is crucial for CNN-based facial landmark localisation.
However, mainly the L2 loss has been used in existing deep-neural-network-based facial landmarking systems.
In this paper, to the best of our knowledge, we are the first to analyse different loss functions for CNN-based facial landmark localisation and demonstrate that the L1 and smooth L1 loss functions perform much better than the L2 loss.
Motivated by our analysis, we propose a new loss function, namely Wing loss, which further improves the accuracy of CNN-based facial landmark localisation systems.
\begin{figure}[t]
\centering
 \includegraphics[trim = 60mm 122mm 58mm 123mm, clip, width = .85\linewidth]{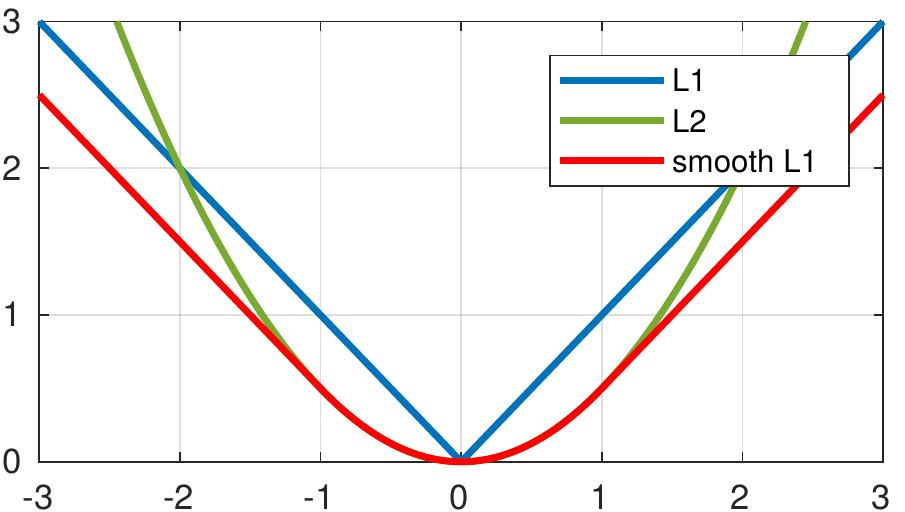}
\caption{Plots of the L1, L2 and smooth L1 loss functions.}
\label{fig_loss}
\end{figure}

\subsection{Analysis of different loss functions}
\label{sec_4_1}
Given a training image $\mathcal{I}$ and a network $\Phi$, we can predict the facial landmarks as a vector $\mathbf{s}' = \Phi (\mathcal{I})$.
The loss is defined as:
\begin{equation}
loss(\mathbf{s}, \mathbf{s}') = \sum_{i=1}^{2L} f(s_i - s'_i),
\label{equ_loss}
\end{equation}
where $\mathbf{s}$ is the ground-truth shape vector of the facial landmarks.
For $f(x)$ in the above equation, L1 loss uses $L1(x) = |x|$ and L2 loss uses $L2(x) = \frac{1}{2}x^2$.
The smooth L1 loss function is piecewise-defined as:
\begin{equation}
smooth_{L1}(x) = \left\{
\begin{array}{ll}
\frac{1}{2}x^2  & \text{if } |x| < 1 \\
|x| - \frac{1}{2}      & \text{otherwise}
\end{array}
\right.,
\end{equation}
which is quadratic for small values of $|x|$ and linear for large values~\cite{girshick2015fast}.
More specifically, smooth L1 uses $L2(x)$ for $x\in(-1,1)$ and shifted $L1(x)$ elsewhere.
Fig.~\ref{fig_loss} depicts the plots of these loss functions.
It should be noted that the smooth L1 loss is a special case of the Huber loss~\cite{huber1964robust}.
The loss function that has widely been used in facial landmark localisation is the L2 loss function.
However, it is well-known that the L2 loss is sensitive to outliers.
This is the main reason why, \eg, Girshick~\cite{girshick2015fast} and Rashid \etal~\cite{ Rashid_2017_CVPR} use the smooth L1 loss function for their localisation tasks.

For evaluation, the AFLW-Full protocol has been used~\cite{zhu2016unconstrained}\footnote{The AFLW dataset is introduced in Section~\ref{sec_7_2_1}.}.
This protocol consists of 20k training images and 4386 test images. Each image has 19 facial landmarks.
We use three state-of-the-art algorithms~\cite{zhu2016unconstrained,Feng_2017_CVPR, Lv_2017_CVPR} as our baseline for comparison.
The first one is the Cascaded Compositional Learning algorithm (CCL)~\cite{zhu2016unconstrained}, which is a multi-view cascaded regression model based on random forests.
The second one is the Two-stage Re-initialisation Deep Regression Network (TR-DRN)~\cite{Lv_2017_CVPR}.
The last baseline algorithm is a multi-view approach based on cascaded shape regression, namely DAC-CSR~\cite{Feng_2017_CVPR}.
\begin{figure}[t]
\centering
 \includegraphics[trim = 25mm 99mm 29mm 105mm, clip, width = .99\linewidth]{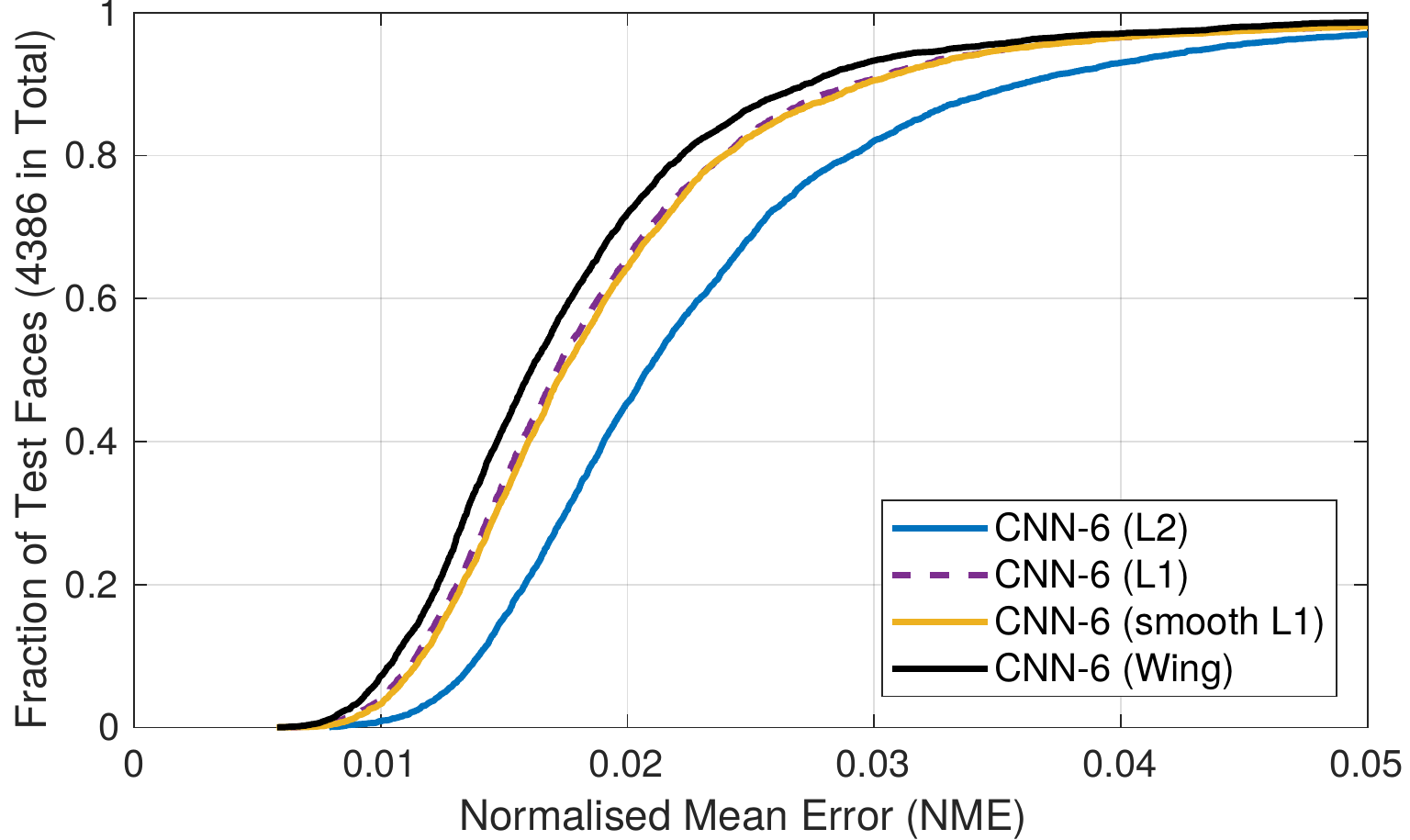}
\caption{CED curves comparing different loss functions on the AFLW dataset, using the AFLW-Full protocol.}
\label{fig_cnn64_aflw_wing}
\end{figure}
\begin{table}[t]
\renewcommand{\arraystretch}{1.1}
\centering
\caption{A comparison of different loss functions with the three baseline algorithms in terms of the average error normalised by face size. Each training has been performed for 120k iterations. The learning rate is reduced from $3\times10^{-6}$ to $3\times10^{-8}$ for L2, and from $3\times10^{-5}$ to $3\times10^{-7}$ for the other loss functions.}
\vspace{3pt}
\small
\label{table1}
\begin{tabular}{lc}
\hline
method & average normalised error \\ 
\hline
CCL (CVPR2016)~\cite{zhu2016unconstrained} & 2.72$\times10^{-2}$ \\
DAC-CSR (CVPR2017)~\cite{Feng_2017_CVPR} & 2.27$\times10^{-2}$ \\
TR-DRN (CVPR2017)~\cite{Lv_2017_CVPR} & 2.17$\times10^{-2}$ \\
\hline
CNN-6 (L2) & 2.41$\times10^{-2}$  \\
CNN-6 (L1) & 2.00$\times10^{-2}$  \\
CNN-6 (smooth L1) & 2.02$\times10^{-2}$  \\
CNN-6 (Wing loss) & 1.88$\times10^{-2}$  \\
\hline
\end{tabular}
\end{table}

We train the CNN-6 network on AFLW using three different loss functions and report the results in Table ~\ref{table1}.
The L2 loss function, which has been widely used for facial landmark localisation, performs well.
The result is better than CCL in terms accuracy but worse than DAC-CSR and TR-DRN.
Surprisingly, when we use L1 or smooth L1 for the CNN-6 training, the performance in terms of accuracy improves significantly and outperforms all the state-of-the-art baseline approaches, despite the CNN network's simplicity.

%%% Wing Loss
%%%------------------------------------------------------------
%%%------------------------------------------------------------
%%%------------------------------------------------------------
%%%------------------------------------------------------------
%%%------------------------------------------------------------
%%%------------------------------------------------------------

\subsection{The proposed Wing loss}
\label{sec_4_2}
We compare the results obtained on the AFLW dataset using the simple CNN-6 network in Fig.~\ref{fig_cnn64_aflw_wing} by plotting the Cumulative Error Distribution (CED) curves.
We can see that all the loss functions analysed in the last section perform well for large errors.
This indicates that the training of a neural network should pay more attention to the samples with small or medium range errors.
To achieve this target, we propose a new loss function, namely Wing loss, for CNN-based facial landmark localisation.

In order to motivate the new loss function, we provide an intuitive analysis of the properties of the L1 and L2 loss functions (Fig.~\ref{fig_loss}). The magnitude of the gradients of these two functions is $1$ and $|x|$ respectively, and the magnitude of the corresponding optimal step sizes should be $|x|$ and $1$. Finding the minimum in either case is straightforward. However, the situation becomes more complicated when we try to optimise simultaneously the location of multiple points, as in our problem of facial landmark localisation formulated in Eq.~(\ref{equ_loss}). In both cases the update towards the solution will be dominated by larger errors. In the case of L1, the magnitude of the gradient is the same for all the points, but the step size is disproportionately influenced by larger errors. For L2, the step size is the same but the gradient will be dominated by large errors. Thus in both cases it is hard to correct relatively small displacements.

The influence of small errors can be enhanced by an alternative loss function, such as $\ln x$. Its gradient, given by $1/x$, increases as we approach zero error. 
The magnitude of the optimal step size is $x^2$. When compounding the contributions from multiple points, the gradient will be dominated by small errors, but the step size by larger errors. This restores the balance between the influence of errors of different sizes. However, to prevent making large update steps in a potentially wrong direction, it is important not to overcompensate the influence of small localisation errors. This can be achieved by opting for a log function with a positive offset. 

This type of loss function shape is appropriate for dealing with relatively small localisation errors. However, in facial landmark detection of in-the-wild faces we may be dealing with extreme poses where initially the localisation errors can be very large. In such a regime the loss function should promote a fast recovery from these large errors. This suggests that the loss function should behave more like L1 or L2. As L2 is sensitive to outliers, we favour L1.

The above intuitive argument points to a loss function which for small errors should behave as a log function with an offset, and for larger errors as L1. Such a composite loss function can be defined as:
\begin{equation}
\label{equ_wing_loss}
wing(x) = \left\{
\begin{array}{ll}
w \ln (1 + |x|/\epsilon)  & \text{if } |x| < w \\
|x| - C  & \text{otherwise}
\end{array}
\right.,
\end{equation}
where the non-negative $w$ sets the range of the nonlinear part to $(-w, w)$, $\epsilon$ limits the curvature of the nonlinear region and $ C = w - w\ln ({1 + w/\epsilon})$ is a constant that smoothly links the piecewise-defined linear and nonlinear parts.
Note that we should not set $\epsilon$ to a very small value because it makes the training of a network very unstable and causes the exploding gradient problem for very small errors.
In fact, the nonlinear part of our Wing loss function just simply takes the curve of $\ln(x)$ between $[\epsilon/w, 1+\epsilon/w)$ and scales it along both the X-axis and Y-axis by a factor of $w$.
Also, we apply translation along the Y-axis to allow $wing(0)=0$ and to impose continuity on the loss function.
\begin{table}[]
\centering
\caption{A comparison of different parameter settings ($w$ and $\epsilon$) for the proposed Wing loss function, measured in terms of the average normalised error ($\times 10^{-2}$) on AFLW using our CNN-6 network.}
\label{table_wing}
\vspace{3pt}
\small
\begin{tabular}{c|cccccc}
\hline
\diagbox{$\epsilon$}{$w$}  & 4 & 6 & 8 & 10 & 12 & 14\\
\hline
0.5 & 1.95  & 1.92  & 1.92  & 1.94 & 1.97 & 1.94\\
1   & 1.95  & 1.91  & 1.91  & 1.90 & 1.90 & 1.95\\
2   & 1.98  & 1.92  & 1.91  & \textbf{1.88} & 1.90 & 1.98\\
3   & 2.02  & 1.96  & 1.93  & 1.91 & 1.89 & 2.02\\
\hline
\end{tabular}
\end{table}

From Fig.~\ref{fig_cnn64_aflw_wing}, we can see that our Wing loss outperforms L2, L1 and smooth L1 in terms of accuracy.
The Wing loss further reduces the average normalised error from $2 \times 10^{-2}$ to $1.88 \times 10^{-2}$, which is $6\%$ lower than the best result obtained in the last section (Table~\ref{table1}) and $13\%$ lower than the best state-of-the-art deep-learning baseline approach, \ie TR-DRN.
In our experiments, we set the parameters of the Wing loss as $w = 10$ and $\epsilon = 2$.
For the results of different parameter settings, please refer to Table~\ref{table_wing}. 

%%% PDB
%%%------------------------------------------------------------
%%%------------------------------------------------------------
%%%------------------------------------------------------------
%%%------------------------------------------------------------
%%%------------------------------------------------------------
%%%------------------------------------------------------------
\section{Pose-based data balancing}
\label{sec_5}
Extreme pose variations are very challenging for robust facial landmark localisation in the wild.
To mitigate this issue, we propose a simple but very effective Pose-based Data Balancing (PDB) strategy.
We argue that the difficulty for accurately localising faces with large poses is mainly due to data imbalance, which is a well-known problem in many computer vision applications~\cite{shrivastava2016training}.
For example, given a training dataset, most samples in it are likely to be near-frontal faces.
The neural network trained on such a dataset is dominated by frontal faces. By over-fitting to the frontal pose it cannot adapt well to faces with large poses.
In fact, the difficulty of training and testing on merely frontal faces should be similar to that on profile faces.
This is the main reason why a view-based face analysis algorithm usually works well for pose-varying faces.
As an evidence, even the classical view-based Active Appearance Model can localise faces with large poses very well (up to $90^{\circ}$ in yaw)~\cite{Cootes2000}.
\begin{figure}[!t]
\centering

\includegraphics[trim = 38mm 125mm 37.5mm 127.5mm, clip, width = 1\linewidth]{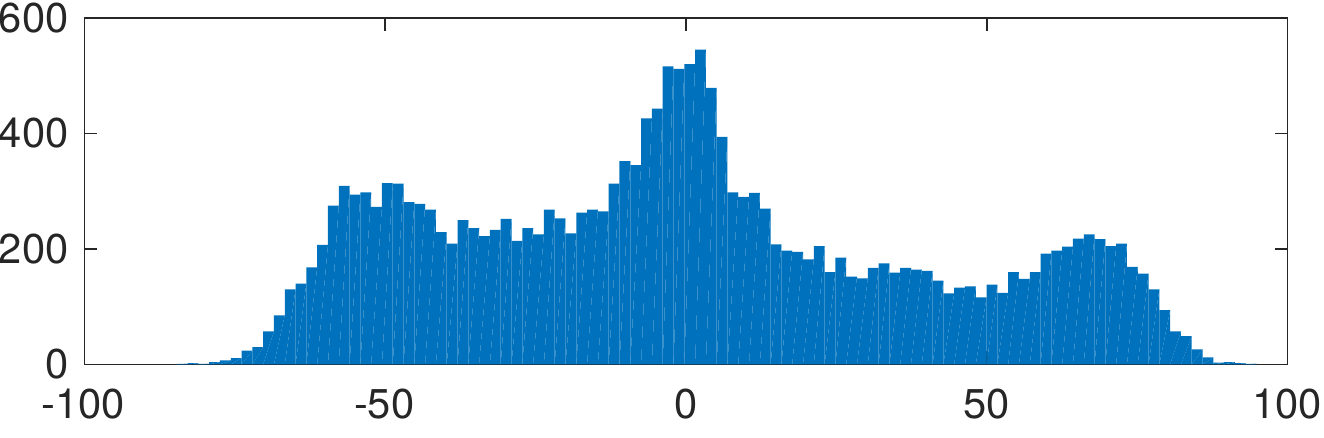}

\includegraphics[trim = 0mm 0mm 0mm 0mm, clip, width = 1\linewidth]{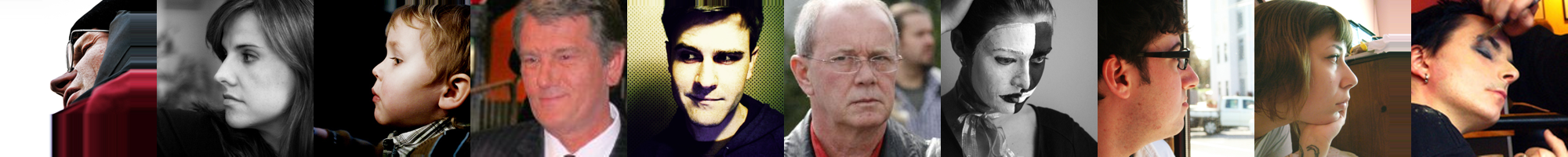}

\caption{Distribution of the pose coefficients of the AFLW training samples by projecting their shapes to the 1-D pose space.}
\label{fig_aflw_pose}
\end{figure}

To perform PDB, we first align all the training shapes to a reference shape using Procrustes Analysis, with the mean shape as the reference shape.
Then we apply PCA to the aligned training shapes and project the original shapes to the one dimensional space defined by the shape eigenvector (pose space) controlling pose variations.
The distribution of projection  coefficient of the training samples is represented by a histogram with $K$ bins, plotted in Figure \ref{fig_aflw_pose}.
With this histogram, we balance the training data by duplicating the samples falling into the bins of lower occupancy. We modify each duplicated sample by performing random image rotation, bounding box perturbation and other data augmentation approaches introduced in Section~\ref{sec_7_1}.
To deal with in-plane rotations, we use a two-stage facial landmark localisation framework that will be introduced in Section~\ref{sec_6}.
The results obtained by the CNN-6 network with PDB are shown in Table~\ref{table_PDB}.
It should be noted that PDB improves the performance of CNN-6 on the AFLW dataset for all different types of loss functions.
\begin{table}[!t]
\centering
\caption{A comparison of different loss functions using our PDB strategy and two-stage landmark localisation framework, measured in terms of the average normalised error ($\times 10^{-2}$) on AFLW. The method CNN-6/7 indicates the proposed two-stage localisation framework using CNN-6 as the first network and CNN-7 as the second network (Section~\ref{sec_6}). For CNN-7, the learning rate is reduced from $1\times10^{-6}$ to $1\times10^{-8}$ for L2, and from $1\times10^{-5}$ to $1\times10^{-7}$ for the L1, smooth L1 and Wing loss functions.}
\vspace{3pt}
\small
\label{table_PDB}
\begin{tabular}{l|cccc}
\hline
\diagbox{method}{loss}   & L2 & L1 & smooth L1 & Wing \\
\hline
CNN-6              &  2.41 & 2.00  &  2.02  &  1.88  \\
CNN-6 + PDB        &  2.23 & 1.89  &  1.91  &  1.83  \\
\hline
CNN-6/7       & 2.06  &  1.82 &  1.84  & 1.71  \\
CNN-6/7 + PDB & 1.94  &  1.73 &  1.76  & 1.65  \\
\hline
\end{tabular}
\end{table}

%%% Multi-Stage CNN + PDB
%%%------------------------------------------------------------
%%%------------------------------------------------------------
%%%------------------------------------------------------------
%%%------------------------------------------------------------
%%%------------------------------------------------------------
%%%------------------------------------------------------------
\section{Two-stage landmark localisation}
\label{sec_6}
Besides the out-of-plane head rotations, the accuracy of a facial landmark localisation algorithm can be degraded by other factors, such as in-plane head rotations and inaccurate bounding boxes output from a poor face detector.
%To address these issues, we can stack or cascade multiple networks to form a coarse-to-fine structure.
%In principle, more complicated networks can be used for performance boosting. 
%However, cascading multiple networks may greatly slow down the speed of the algorithm both for the offline training and online testing stages.
To mitigate this issue, we advocate the use of a two-stage landmark localisation framework.

In the proposed two-stage localisation framework, we use a very simple network, \ie the CNN-6 network with $64\times64\times3$ input images, as the first network.
The CNN-6 network is very fast (400 fps on an NVIDIA GeForce GTX Titan X \textit{Pascal}), hence it will not slow down the speed of our facial landmark localisation algorithm too much.
The landmarks output by the CNN-6 network are used to refine the input image for the second network by removing the in-plane head rotation and correcting the bounding box.
Also, the input image resolution for the second network is increased for fine-grained landmark localisation from $64 \times 64 \times3$ to $128 \times 128 \times 3$, with the addition of one set of convolutional, Relu and Max pooling layers. Hence, the term `CNN-7' is used to denote the second network.
The CNN-7 network has a similar architecture to the CNN-6 network in Fig.~\ref{fig_cnn64}.
The difference is that CNN-7 has 6 convolutional layers which resize the feature map from $128 \times 128 \times3$ to $2 \times 2 \times 512$.
In addition, for the first convolutional layer in CNN-7, we double the number of $3 \times 3$ kernels from $32$ to $64$.
We use the term `CNN-6/7' for our two-stage facial landmark localisation framework and compare it with the CNN-6 network in Table~\ref{table_PDB}.
As reported in the table, the use of our two-stage landmark localisation framework further improves the accuracy, regardless of the type of loss function used.

%%% Results
%%%------------------------------------------------------------
%%%------------------------------------------------------------
%%%------------------------------------------------------------
%%%------------------------------------------------------------
%%%------------------------------------------------------------
%%%------------------------------------------------------------
\section{Experimental results}
\label{sec_7}
In this section, we evaluate our method on the Annotated Facial Landmarks in the Wild (AFLW) dataset~\cite{aflw_dataset_2011} and the 300 Faces in the Wild (300W) dataset~\cite{sagonas2013300}.
We first introduce our implementation details and experimental settings.
Then we compare our algorithm with state-of-the-art approaches on AFLW and 300W.
Last, we analyse the performance of different networks in terms of both accuracy and speed.

\subsection{Implementation details}
\label{sec_7_1}
In our experiments, we used Matlab 2017a and the MatConvNet toolbox\footnote{http://www.vlfeat.org/matconvnet/}.
The training and testing of our networks were conducted on a server running Ubuntu 16.04 with $2\times$ Intel Xeon E5-2667 v4 CPU, 256 GB RAM and 4 NVIDIA GeForce GTX Titan X (\textit{Pascal}) cards.
Note that we only use one GPU card for measuring the run time.
We set the weight decay to $5\times10^{-4}$, momentum to $0.9$ and batch size to $8$ for network training.
Each model was trained for 120k iterations.
We did not use any other advanced techniques in our CNN-6 and CNN-7 networks, such as batch normalisation, dropout or residual blocks.
The standard ReLu function was used for nonlinear activation, and Max pooling with the stride of $2$ was used to downsize feature maps.
For the convolutional layer, we used $3\times3$ kernels with the stride of $1$.
All our networks, except ResNet-50, were trained from scratch without any pre-training on any other dataset.
For the proposed PDB strategy, the number of bins $K$ was set to $17$ for AFLW and $9$ for 300W.

For CNN-6, the input image size is $64 \times 64 \times 3$.
We reduced the learning rate from $3\times10^{-6}$ to $3\times10^{-8}$ for the L2 loss, and from $3\times10^{-5}$ to $3\times10^{-7}$ for the other loss functions.
The parameters of the Wing loss were set to $w = 10$ and $\epsilon = 2$.
For CNN-7, the input image size is $128 \times 128 \times 3$.
We reduced the learning rate from $1\times10^{-6}$ to $1\times10^{-8}$ for the L2 loss, and from $1\times10^{-5}$ to $1\times10^{-7}$ for the other loss functions.
The parameters of the Wing loss were set to $w = 15$ and $\epsilon = 3$.

To perform data augmentation, we randomly rotated each training image between $[-30, 30]$ degrees for CNN-6 and between $[-10, 10]$ degrees for CNN-7.
In addition, we randomly flipped each training image with the probability of $50\%$.
For bounding box perturbation, we applied random translations to the upper-left and bottom-right corners of the face bounding box within $5\%$ of the bounding box size.
Last, we randomly injected Gaussian blur ($\sigma = 1$) to each training image with the probability of $50\%$.

\textbf{Evaluation Metric:} For evaluation of a facial landmark localisation algorithm, we adopted the widely used Normalised Mean Error (NME).
%Specifically, given the ground truth shape $\mathbf{s}$ and the predicted shape $\mathbf{s}'$ of a test image, the NME is defined as:
%\begin{equation}
%NME = \frac{1}{L} \sum_{i=1}^{L} \frac{\sqrt{(x_l - x'_l)^{2} + (y_l - y'_l)^2}}{d},
%\end{equation}
%where $L$ is the number of facial landmarks, $(x_l, y_l)$ is the coordinates of the $l$th facial landmark and $d$ is the normalisation term.
For the AFLW dataset using the AFLW-Full protocol, the given face bounding box of a test sample is a square~\cite{zhu2016unconstrained}.
To calculate the NME of a test sample, the AFLW-Full protocol uses the width (or height) of the face bounding box as the normalisation term.
For the 300W dataset, we followed the protocol used in~\cite{ren2016face}.
This protocol uses the inter-pupil distance as the normalisation term, which is different from the standard 300W protocol that uses the outer eye corner distance.

\subsection{Comparison with state of the art}
\subsubsection{AFLW}
\label{sec_7_2_1}
We first evaluated our algorithm on the AFLW dataset~\cite{aflw_dataset_2011}, using the AFLW-Full protocol~\cite{zhu2016unconstrained}.
AFLW is a very challenging dataset that has been widely used for benchmarking facial landmark localisation algorithms.
The images in AFLW consist of a wide range of pose variations in yaw (from $-90^\circ$ to $90^\circ$), as shown in Fig.~\ref{fig_aflw_pose}.
The AFLW-Full protocol contains 20,000 training and 4,386 test images, and each image has 19 manually annotated facial landmarks.
\begin{figure}[t]
\centering
 \includegraphics[trim = 17mm 80mm 18mm 90mm, clip, width = 1\linewidth]{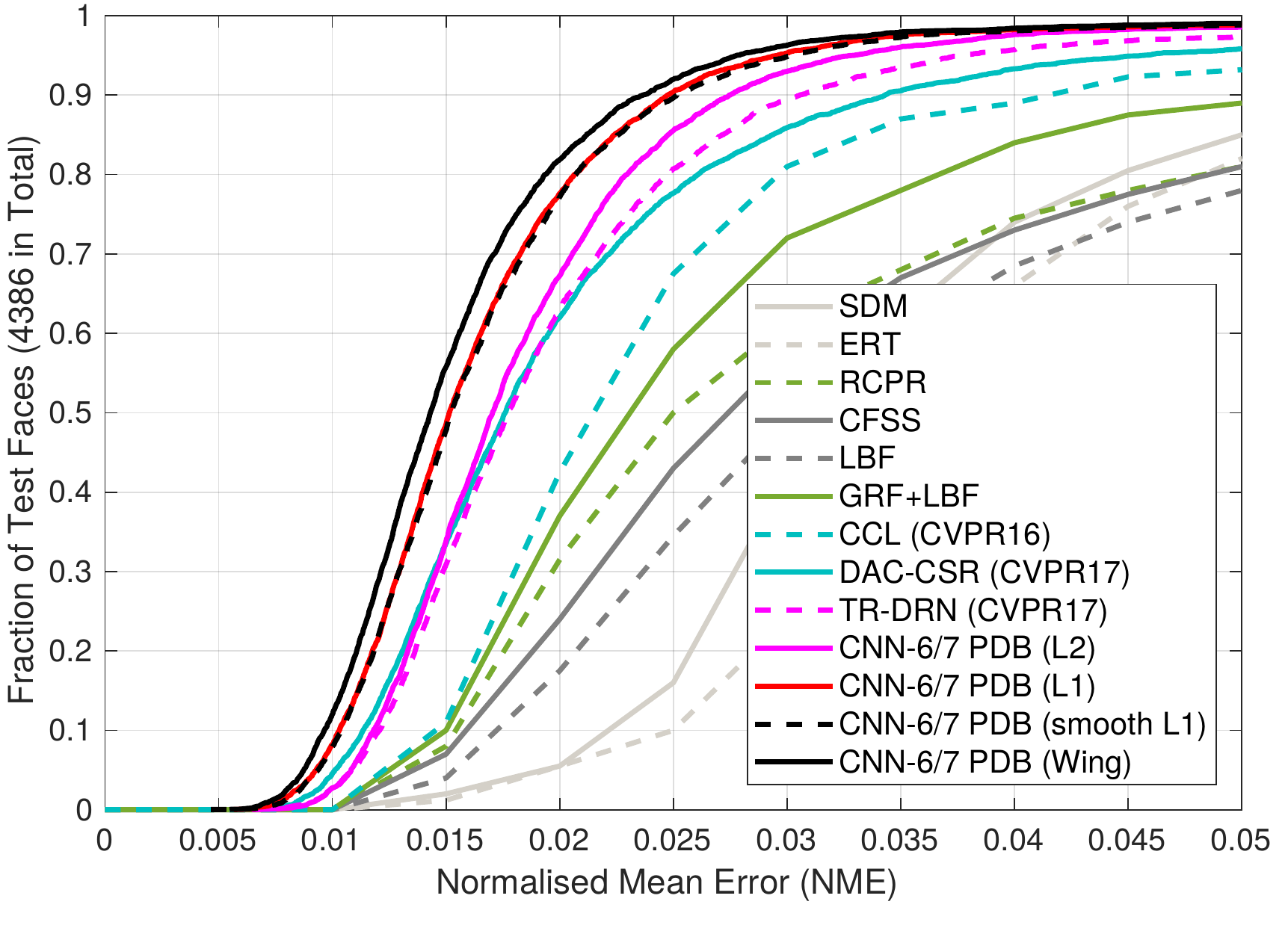}
\caption{A comparison of the CED curves on the AFLW dataset. We compare our method with a set of state-of-the-art approaches, including SDM~\cite{xiong2013supervised}, ERT~\cite{kazemi2014one}, RCPR~\cite{Burgos-Artizzu2013}, CFSS~\cite{zhu2015face}, LBF~\cite{ren2016face}, GRF~\cite{hara2014growing}, CCL~\cite{zhu2016unconstrained}, DAC-CSR~\cite{Feng_2017_CVPR} and TR-DRN~\cite{Lv_2017_CVPR}.}
\label{fig_aflw_final}
\end{figure}

We compare the proposed method with state-of-the-art approaches in terms of accuracy in Fig.~\ref{fig_aflw_final} using the Cumulative Error Distribution (CED) curve.
In our experiments, we used our two-stage facial landmark localisation framework by stacking the CNN-6 and CNN-7 networks (denoted by CNN-6/7), as introduced in Section~\ref{sec_6}.
In addition, the proposed Pose-based Data Balancing (PDB) strategy was adopted, as presented in Section~\ref{sec_5}.
We report the results of the proposed approach using four different loss functions.

As shown in Fig.~\ref{fig_aflw_final}, our CNN-6/7 network outperforms all the other approaches even when trained with the commonly used L2 loss function (magenta solid line).
This validates the effectiveness of the proposed two-stage localisation framework and the PDB strategy.
Second, by simply switching the loss function from L2 to L1 or smooth L1, the performance of our method has been improved significantly (red solid and black dashed lines).
Last, the use of our newly proposed Wing loss function further improves the accuracy (black solid line).
The proportion of test samples (Y-axis) associated with a small to medium normalised mean error (X-axis) is increased.
\begin{table}[!t]
\centering
\caption{A comparison of the proposed approach with the state-of-the-art approaches on the 300W dataset in terms of the NME averaged over all the test samples. We follow the protocol used in~\cite{ren2016face}. Note that the error is normalised by the inter-pupil distance, rather than the outer eye corner distance.}
\vspace{3pt}
\footnotesize
\label{table_300w}
\begin{tabular}{l|ccc}
\hline
\diagbox{method}{subset}   & Com. & Challenge & Full \\
\hline
RCPR~\cite{Burgos-Artizzu2013}   & 6.18 & 17.26 & 8.35 \\
CFAN~\cite{zhang2014coarse}   & 5.50 & 16.78 & 7.69 \\
ESR~\cite{Cao2014face}    & 5.28 & 17.00 & 7.58 \\
SDM~\cite{xiong2013supervised}    & 5.60 & 15.40 & 7.52 \\
ERT~\cite{kazemi2014one}    & -    & -     & 6.40 \\
CFSS~\cite{zhu2015face}   & 4.73 & 9.98  & 5.76 \\
TCDCN~\cite{zhang2014facial}  & 4.80  & 8.60  & 5.54 \\
LBF~\cite{ren2016face}    & 4.95 & 11.98 & 6.32\\
3DDFA~\cite{zhu2016face} & 6.15 & 10.59 & 7.01 \\
3DDFA + SDM & 5.53 & 9.56 & 6.31 \\
DDN~\cite{yu2016deep}    & -    & -     & 5.65 \\
RAR~\cite{Xiao2016} & 4.12 & 8.35 & 4.94 \\
DeFA~\cite{Liu_2017_ICCV_Workshops} & 5.37 & 9.38 & 6.10 \\
TR-DRN~\cite{Lv_2017_CVPR} & 4.36 & 7.56  & 4.99 \\
RCN~\cite{honari2016recombinator} & 4.70 & 9.00 & 5.54 \\
RCN$^+$~\cite{honari2018improving} & 4.20 & 7.78 & 4.90 \\
\hline
CNN-6/7 + PDB (L2) & 4.18  & 8.19  & 4.97  \\
CNN-6/7 + PDB (L1) & 3.58 & \textbf{7.02}  & 4.26  \\
CNN-6/7 + PDB (smooth L1) & 3.57  & 7.08  & 4.26  \\
CNN-6/7 + PDB (Wing) & \textbf{3.27}  &  7.18 &  \textbf{4.04}    \\
\hline
%ResNet-50 + PDB (Wing) & \textbf{3.01} & \textbf{6.01}  & \textbf{3.60}\\
%ResNet-152 + PDB (Wing) & 3.09  & 6.22  & 3.70 \\
%\hline
\end{tabular}
\end{table}

\subsubsection{300W}
The 300W dataset is a collection of multiple face datasets, including LFPW~\cite{Belhumeur2011}, HELEN~\cite{le2012interactive}, AFW~\cite{zhu2012face} and XM2VTS~\cite{Messer1999}.
The face images involved in 300W have been semi-automatically annotated by 68 facial landmarks~\cite{sagonas2013semi}.
To perform the evaluation on 300W, we followed the protocol used in~\cite{ren2016face}.
The protocol uses the full set of AFW and the training subsets of LFPW and HELEN as the training set, which contains 3148 training samples in total.
The test set of the protocol includes the test subsets of LFPW and HELEN, as well as 135 IBUG face images newly collected by the managers of the 300W dataset.
The final size of the test set is 689.
The test set is further divided into two subsets for evaluation, \ie the common and challenging subsets.
The common subset has 554 face images from the LFPW and HELEN test subsets and the challenging subset constitutes the 135 IBUG face images.

Similar to the experiments conducted on the AFLW dataset, we used the two-stage localisation framework with our PDB strategy.
The results obtained by our approach with different loss functions are reported in Table~\ref{table_300w}.

As shown in Table~\ref{table_300w}, our two-stage landmark localisation framework with the PDB strategy and the newly proposed Wing loss function outperforms all the other state-of-the-art algorithms on the 300W dataset in accuracy.
The error has been reduced by almost $20\%$ as compared to the current best result reported by the RAR algorithm~\cite{Xiao2016}.

\subsection{Run time and network architectures}
Facial landmark localisation has been widely used in many real-time practical applications, hence the speed together with accuracy of an algorithm is crucial for the deployment of the algorithm in commercial use cases.

To analyse the performance of our Wing loss on more advanced network architectures, we evaluated ResNet~\cite{he2016deep} for the task of landmark localisation on AFLW and 300W. 
We used the ResNet-50 model that was pre-trained on the ImageNet ILSVRC classification problem\footnote{http://www.vlfeat.org/matconvnet/pretrained/}.
We fine-tuned the model on the training sets of AFLW and 300W separately for landmark localisation.
The input for ResNet is a $224 \times 224 \times 3$ colour image.
It should be highlighted that, to our best knowledge, this is the first time that such a deep network has been used for facial landmark localisation.

For both AFLW and 300W, by replacing the CNN-6/7 network with ResNet-50, the performance has been further improved by around 10\%, as shown in Table~\ref{table_resnet}.
However, this performance boosting comes at the cost of much slower training and inference of ResNet compared to CNN-6/7.
\begin{table}[t]
\centering
\caption{A comparison of our simple network with ResNet-50, in terms of accuracy on AFLW-Full and 300W.}
\vspace{3pt}
\footnotesize
\label{table_resnet}
\begin{tabular}{l|c|ccc}
\hline
 \multirow{2}{*}{} & \multirow{2}{*}{AFLW} & \multicolumn{2}{c}{300W} \\
 \cline{3-5}
        &           &   Com.  &   Challenge    & Full  \\
\hline
CNN-6 + PDB (Wing)& 1.83 & 3.35 & 7.20 & 4.10\\
CNN-6/7 + PDB (Wing)& 1.65 & 3.27 & 7.18 &4.04 \\
ResNet-50 + PDB (Wing)& 1.47 & 3.01 & 6.01 &3.60 \\
%ResNet-152 & - & 3.09 & 6.22 &3.70 \\
\hline
\end{tabular}
\end{table}

To validate the effectiveness of our Wing loss for large capacity networks, we also conducted experiments using ResNet-50 with different loss functions on AFLW. The results are reported in Table.~\ref{table_resnet2}. The results further demonstrate the superiority of the proposed Wing loss over other loss functions for large capacity networks, \eg ResNet-50.
\begin{table}[t]
\centering
\caption{A comparison in accuracy of ResNet-50 using different loss functions, evaluated on AFLW-Full.}
\footnotesize
\label{table_resnet2}
\begin{tabular}{l|cccc}
\hline
Loss Function& L2 & L1 & smooth L1 & Wing \\
\hline
NME ($\times10^{-2}$) & 1.68 & 1.51  & 1.52 & 1.47 \\
\hline
\end{tabular}
\end{table}

Last, we evaluated the speed of different networks on the 300W dataset with 68 landmarks for both GPU and CPU devices.
The results are reported in Table~\ref{table_speed}. 
According to the table, our simple CNN-6/7 network is roughly an order of magnitude faster than ResNet-50 at the compromise of 10\% performance difference in accuracy.
Also, our CNN-6/7 model is much faster than most existing DNN-based facial landmark localisation approaches such as TR-DRN~\cite{Lv_2017_CVPR}.
The speed of TR-DRN is 83 fps on an NVIDIA GeForce~GTX Titan~X card.
Even with a powerful GPU card, it is hard to achieve video rate (60fps) with ResNet-50.
It should be noted that our CNN-6/7 still outperforms the state-of-the-art approaches by a significant margin while running at 170 fps on a GPU card, as shown in Fig.~\ref{fig_aflw_final}. 
\begin{table}[t]
\centering
\caption{A comparison of different networks, in the number of model parameters, model size and speed.}
\vspace{3pt}
\footnotesize
\label{table_speed}
\begin{tabular}{l|c|c|cc}
\hline
\multirow{2}{*}{network} & \multirow{2}{*}{\# params} & \multirow{2}{*}{size} & \multicolumn{2}{c}{speed (fps)} \\
\cline{4-5}
        &          &    &   GPU   &   CPU      \\
\hline
CNN-6 & 3.8 M & 14 MB & 400  &  150 \\
%CNN-7 & 8.5 M & 32 MB & 300 &   20 \\
CNN-6/7 & 12.3 M & 46 MB & 170 &   20 \\
ResNet-50 & 25 M & 99 MB & 30 & 8 \\
%ResNet-152 & 60 M& 231 MB & 7 & 2 \\
\hline
\end{tabular}
\end{table}

%\subsubsection{COFW}
%Caltech Occluded Faces in the Wild (COFW) dataset~\cite{Burgos-Artizzu2013}.The COFW dataset has 1345 training and 507 test images, which are all unconstrained faces. Each COFW face has 29 manually annotated landmarks. COFW is a challenging benchmark containing major occlusions.

%%% Conclusion
%%%------------------------------------------------------------
%%%------------------------------------------------------------
%%%------------------------------------------------------------
%%%------------------------------------------------------------
%%%------------------------------------------------------------
%%%------------------------------------------------------------
\section{Conclusion}
\label{sec_8}
In this paper, we analysed different loss functions that can be used for the task of regression-based facial landmark localisation.
We found that L1 and smooth L1 loss functions perform much better in accuracy than the L2 loss function.
Motivated by our analysis of these loss functions, we proposed a new, Wing loss performance measure.
The key idea of the Wing loss criterion is to increase the contribution of the samples with small and medium size errors to the training of the regression network.
To prove the effectiveness of the proposed Wing loss function, extensive experiments have been conducted using several CNN network architectures.
Furthermore, a pose-based data balancing strategy and a two-stage landmark localisation framework were advocated to improve the accuracy of CNN-based facial landmark localisation further.
By evaluating our algorithm on multiple well-known benchmarking datasets, we demonstrated the merits of the proposed approach.

It should be emphasised that the proposed Wing loss is relevant to other regression-based computer vision tasks using convolutional neural networks. However, being constrained by the space limitations, we leave the discussion of its extended use to future reports.

\section*{Acknowledgements}
This work was supported in part by the EPSRC Programme Grant (FACER2VM) EP/N007743/1, EPSRC/dstl/MURI project EP/R018456/1,
 the National Natural Science Foundation of China (61373055, 61672265) and the NVIDIA GPU Grant Program.

{\small
\bibliographystyle{ieee}
\bibliography{mybib}
}

\end{document}